\pdfoutput=1

\documentclass[letterpaper, 10 pt, conference]{ieeeconf}  % Comment this line out if you need a4paper

\IEEEoverridecommandlockouts                              % This command is only needed if 
\usepackage{hyperref}
\usepackage{cite}
\usepackage{xcolor}
\usepackage{subcaption}
\usepackage{multirow}
\usepackage[algo2e,ruled,vlined]{algorithm2e}
\usepackage[utf8]{inputenc}
\usepackage[T1]{fontenc}
\usepackage{pifont}
\usepackage{color,soul}
\usepackage{booktabs}

\usepackage{graphics} % for pdf, bitmapped graphics files
\usepackage{epsfig} % for postscript graphics files
\usepackage{amsmath} % assumes amsmath package installed
\usepackage{amssymb}  % assumes amsmath package installed

\title{\LARGE \bf
End-to-end grasping policies for human-in-the-loop robots \\
via deep reinforcement learning*
}

\author{Mohammadreza Sharif$^{1}$, Deniz Erdogmus$^{1}$, Christopher Amato$^{2}$, and Taskin Padir$^{1}$% <-this % stops a space
\thanks{*This work was supported by NSF under the award number 1544895, 1928654, 1935337 and 1944453.}% <-this % stops a space
\thanks{$^{1}$Electrical and Computer Engineering Department,
        Northeastern University, Boston, MA 02115, USA
        {\tt\small \{sharif.mo,deniz,tpadir\}@northeastern.edu }}%
\thanks{$^{2}$Khoury College of Computer Sciences,
        Northeastern University,
        Boston, MA 02115, USA
        {\tt\small c.amato@northeastern.edu }}%
}

\begin{document}

\maketitle
\thispagestyle{empty}
\pagestyle{empty}

%%%%%%%%%%%%%%%%%%%%%%%%%%%%%%%%%%%%%%%%%%%%%%%%%%%%%%%%%%%%%%%%%%%%
\begin{abstract}

State-of-the-art human-in-the-loop robot grasping is hugely suffered by Electromyography (EMG) inference robustness issues. As a workaround, researchers have been looking into integrating EMG with other signals, often in an \textit{ad hoc} manner. In this paper, we are presenting a method for end-to-end training of a policy for human-in-the-loop robot grasping on \textit{real} reaching trajectories. For this purpose we use Reinforcement Learning (RL) and Imitation Learning (IL) in DEXTRON (DEXTerity enviRONment), a stochastic simulation environment with real human trajectories that are augmented and selected using a Monte Carlo (MC) simulation method. We also offer a success model which once trained on the expert policy data and the RL policy roll-out transitions, can provide transparency to how the deep policy works and when it is probably going to fail.

\end{abstract}

%%%%%%%%%%%%%%%%%%%%%%%%%%%%%%%%%%%%%%%%%%%%%%%%%%%%%%%%%%%%%%%%%%%%%

%%%%%%%%%%%%%%%%%%%%%%%%%
%%    INTRODUCTION     %%
%%%%%%%%%%%%%%%%%%%%%%%%%
\section{Introduction}
\label{sec:introduction}

Robot prosthetic hands provide a solution to bring back part of the lost ability of people with upper limb amputation by offering a way to interact with the environment through actuated fingers. Traditionally, there have been several methods to control a robot prosthetic hand, such as Electromyographic (EMG) control, Electroneurographic (ENG) control, and body-powered control methods. Although EMG control arguably provides the finest control by tele-operating the robot actuators, it is not the prevalent method of choice in the market. This lack of popularity of the EMG control method is mainly attributed to its lack of robustness, which is explained by real-life versus lab variations such as electrode shift \cite{Muceli2014-lx}, skin-electrode impedance, muscle fatigue \cite{Hakonen2015-rt}, and stump posture change \cite{Hwang2017-to}. This lack of robustness often causes unpredictability which can lead to passive usage of the robot hand or its total abandonment \cite{Chadwell2016-zk}. Some works have tackled the robustness issue via increasing robot autonomy by using other informative signals along with EMG. Dosen et al. \cite{Dosen2010-tg} et al. use RGB-D camera to detect grasp type. Gigli et al. \cite{Gigli2017-ih} use Gaze information as a guide for detecting human intent and grasp type. In \cite{Sharif2019-sk}, human hand trajectories are utilized to infer time-to-arrive which can be used to gradually shape the hand. Zhuang et al. \cite{Zhuang2019-td} use tactile sensors to grasp the object whenever the hand contacts the object. A low-level controller then maintains the force to avoid object slippage. During reach-to-grasp motion, humans change the shape of their hand gradually to conform to the object contours \cite{Santello1998-xb}. While EMG control provides more intuitive control than the other control methods like body-powered control, due to delayed \cite{Smith2011-fz} and noisy inference, the overall motion kinematics seems less smooth and reach/grasp seem to be less coordinated with respect to healthy humans \cite{Bouwsema2010-me}. This compromises grasp intuitiveness and the overall time for grasping \cite{Bouwsema2010-me}. EMG control method also needs active user attention and excessive energy consumption \cite{Merad2018-ju} which may not be desirable for lasting and repetitive activities. In this work, we do not use EMG signals. Rather, we assume knowledge of the human intent by knowing the task \textit{a priori}.

\begin{figure}[tpb]
        \centering
        \includegraphics[width=0.40\textwidth]{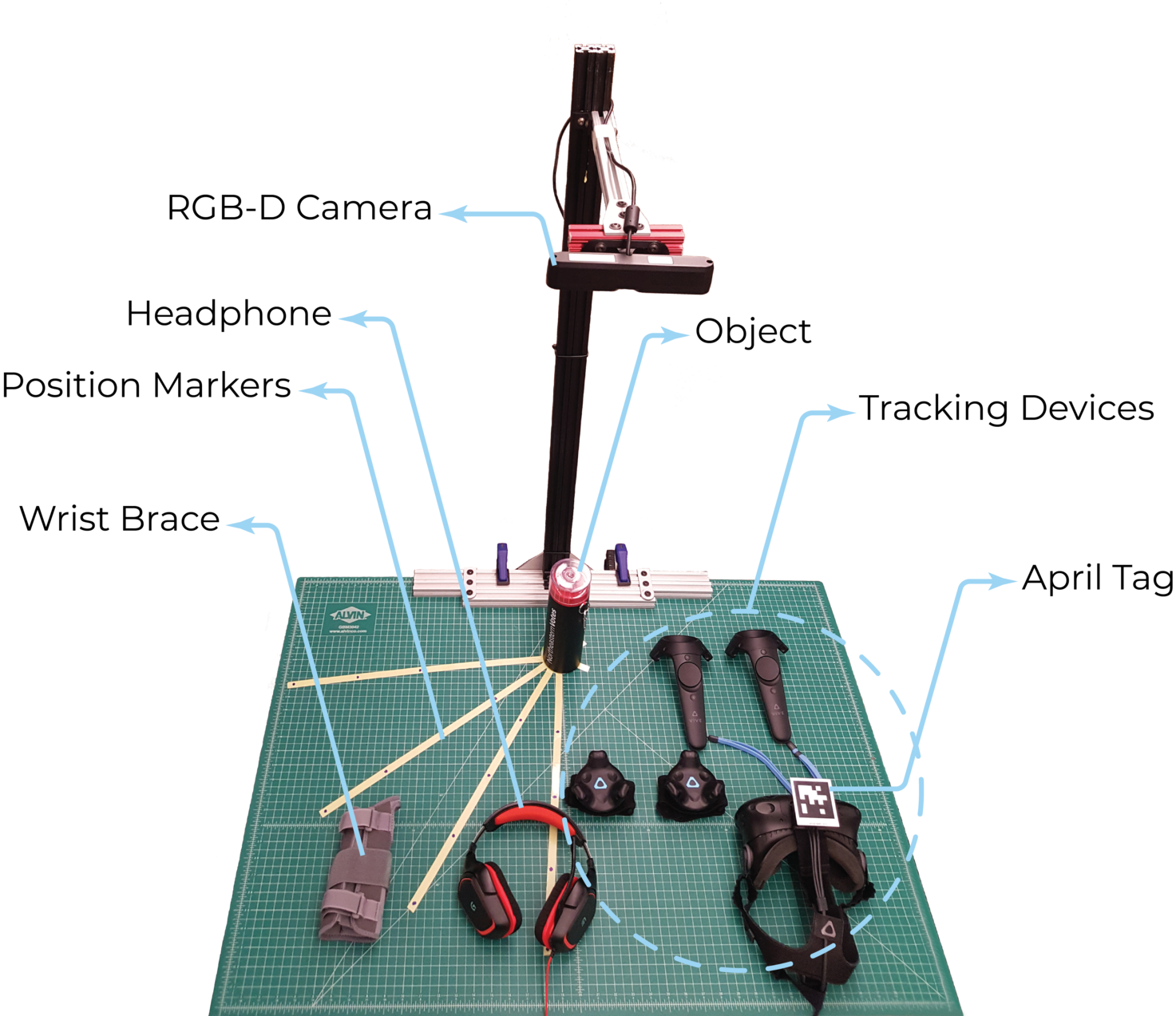}
        \caption{The experimental setup for collecting real human trajectories for DEXTRON.}
        \label{fig:workspace}
\end{figure}

RL is a wide-spread method to address robotic grasping and manipulation problems \cite{Zhang2018-kl, Gu2016-ge}. RL has also been used for robot prosthetic hands control. Pilarski et al. \cite{Pilarski2011-nh} use a continuous actor-critic RL method to learn finger actuation commands from EMG signal inputs. IL is a method that can be used to help RL find optimal solutions in shorter time \cite{Ross2010-br, Gao2018-ql}. Vasan et al. \cite{Vasan2017-og} collect EMG signals from the intact hand as the demonstrator and learn a policy using RL for the robot prosthetic hand at the other side. Ficuciello et al. \cite{Ficuciello2019-wo} leverage IL to learn a policy for grasping objects in the synergy space and then refine the learned policy using RL. RL and IL (RLIL) were used in \cite{Sharif2020-ll} to learn an end-to-end policy for robot prosthetic hands where human hand transport was modeled by minimum-jerk trajectories.

In this paper, we are proposing learning an end-to-end policy for below-elbow upper limb robot prosthetic hands based on \textit{real} human hand reaching trajectories. Knowing the human intent \textit{a priori}, we learn a policy that not only gradually molds intuitively to the object shape, but also maintains the required grasp force on the object without relying on EMG. Unlike \cite{Sharif2020-ll} where simulated hand trajectories were used, we create a stochastic environment with \textit{real} hand reaching trajectories, called DEXTRON (DEXTerity enviRONment), from data we collect from several human subjects. Real trajectories impose several challenges which make the RL and IL techniques used in \cite{Sharif2020-ll} not readily available. By not relying on the EMG signals, our method can also relax some of the other EMG control side-effects like continuous attention requirement, excessive energy consumption, and fatigue.

Contrary to the classic robot manipulation problem where the robot controls both the end-effector transportation and grasp, in the below-elbow robot prosthesis grasping problem the robot can merely control the fingers (and the wrist in some robots). Accordingly, since human and the robot are collaborating to reach a shared goal \cite{Grosz1996-hh}, e.g. grasping an object, this problem is categorized as human-robot collaboration (HRC), regardless of whether a shared control strategy is used for controlling the wrist and fingers or not. In HRC, both human and the robot can adapt to the other side's policy, i.e. human-robot mutual adaptation \cite{Nikolaidis2017-ce}. In this paper, we propose a predictive model, called a success model, which predicts the final outcome of the robot in grasping the object. This model, not only provides a way for more transparency for robot decisions through post hoc explanation \cite{Barredo_Arrieta2020-bw}, but also helps human get adapted to the robot quicker by getting more insight and thus trust into the robot control strategy \cite{Amirshirzad2019-sa, Lewis2018-dq}. This model is inspired by \cite{Levine2016-cq} in which a similar model is used to predict the final expected outcome for grasping if an action is executed at a given state.

The contributions of this paper are: 1) development of an  extension to the RL and IL method proposed in \cite{Sharif2020-ll} to learn end-to-end grasping policy for robot prosthetic hands with real transport trajectories based on Monte Carlo simulation, 2) development of a stochastic simulation environment, DEXTRON, for robot prosthetic hands in Google DeepMind dm\_control \cite{Tassa2018-qp} which can be used by researchers as a stochastic environment with delayed sparse rewards in order to improve results of the current study or solve new problems regarding robot prosthetic hands, and 3) introduction of a success model to obtain insights into the trained policy and to compare it with the demonstration policy qualitatively. The source code of this paper is available on GitHub \footnote{https://github.com/sharif1093/dextron} under BSD open source license.

%%%%%%%%%%%%%%%%%%%%%%%%%
%%       METHODS       %%
%%%%%%%%%%%%%%%%%%%%%%%%%
\section{Methods}
\label{sec:methods}
%%%%%%%%%%%%%%%%%%%%%%%%%%%%%%%%%%%%%%%%%%%%%%%
\subsection{Problem definition and assumptions \label{sub:problem_definition_and_assumptions}}

We have a robot prosthetic hand for below-elbow amputation. The wrist is assumed to be fixed. The intent of the human is to grasp the only object in the simulation environment - a bottle always located at the world frame origin. We assume that the robot knows the human intent \textit{a priori}. We want to learn an end-to-end policy which maps the environment states to the finger actuators' velocity commands. Hand transport trajectories are based on real hand transport trajectories collected from human subjects. A 5 degree of freedom robot prosthetic hand with 5 degrees of actuation is modeled in a simulation environment in agreement with common robot prosthetic hands in the market. We assume that all of our velocity-controlled fingers are coupled together, i.e. they all receive the same command, and our action space is $\mathcal{A} = [-1,1]$, where an action greater than zero indicates a closing command. We further assume we have access to the raw system states $\mathcal{S} \subset \mathbb{R}^{48}$, which includes all joint and link positions and quaternions as well as their time derivatives. As opposed to our previous study \cite{Sharif2020-ll}, we had to change our definition for the reward function in order to accommodate inter-subject variations fairly. After the first timestep the object or the robot hand are above a threshold, a timer starts counting down for 20 timesteps. Then, for every timestep that the object is above the threshold, the agent is rewarded $r=+1$. With this reward function, the maximum sum of rewards an agent can receive in the environment is $+20$, and that happens whenever the agent grasps the object successfully and the object does not slide in the robot hand, showing the force has been maintained on the object. If the robot initially grasps the object successfully but later the object slides in the hand, the sum of rewards will be less than $+20$, and when the agent fails to grasp the object altogether, the sum of rewards would be $0$.

For differentiation between the simulated trajectory environment (introduced in \cite{Sharif2020-ll}) and our new real trajectory environment (introduced in this paper), we call them the simulated-trajectory environment and DEXTRON, respectively. Also, \textit{environment} will refer to DEXTRON unless explicitly indicated otherwise.

%%%%%%%%%%%%%%%%%%%%%%%%%%
\subsection{DEXTRON instantiation}

We define a trajectory, $\lambda$, as a sequence that indicates pose of the human subject's hand over time, i.e. $\lambda = \big\{ (\mathbf{x}_i,\mathbf{q}_i) | i=1,...,T \big\}$, where $\mathbf{x}_i$ is the 3d position of a frame attached to the subject's hand at time instance $i$, $\mathbf{q}_i$ is the frame's quaternion at time instance $i$ with respect to the global frame, and $T$ is the total number of timesteps of the trajectory. We define $\mathbf{\delta}=[\delta_x,\delta_y,0]^T$ as the trajectory offset, and we have $\lambda+\mathbf{\delta} \triangleq  \big\{ (\mathbf{x}_i + \mathbf{\delta},\mathbf{q}_i) | i=1,...,T \big\}$. We define $\Lambda$ as the set of all collected \textit{real} human hand trajectories. In order to further enrich diversity within the set of trajectories in $\Lambda$, we also randomize the duration of each trajectory by scaling its duration to $t_{\text{new}} = t_{\text{old}} \times t_s + t_n$, where $t_{\text{new}}$ is the new duration of the trajectory, $t_{\text{old}}$ is the old duration of the trajectory, $t_s$ is a constant scaling factor, and $t_n$ is the duration offset. We also resample each new trajectory to obtain an evenly sampled trajectory with a sampling interval of $0.02s$. We use piecewise B-spline interpolation for resampling $\mathbf{x}_i$ and Squad method \cite{Dam1998-zj} for resampling $\mathbf{q}_i$.

To instantiate DEXTRON, we sample the environment settings as $\lambda \sim \mathcal{U}(\Lambda)$, $\delta_x \sim \mathcal{U}(\Delta_x)$, $\delta_y \sim \mathcal{U}(\Delta_y)$, and $t_n \sim \mathcal{N}(\mu_t, \sigma_t)$, where $\mathcal{U}(A)$ indicates a uniform distribution over all elements of set $A$. We call the tuple $(\lambda,\delta_x,\delta_y,t_n)$ the environment settings. The sets and other environment parameters are provided in Table~\ref{tab:param_dists}.

%%%%%%%%%%%%%%%%%%%%%%%%%%%%%%%%%%%%%%%%%%%%%%%%
\subsection{Creating the set of expert policies \label{sub:creating_the_set_of_expert_policies}}

In our previous study \cite{Sharif2020-ll}, we used the following two-phase p-controller as the expert policy, which roughly mimicked the human aperture control behavior \cite{Hoff1993-xu}:

\begin{equation} \label{eq:expert_policy_class}
	\pi_e(h, \hat{d}) = \left\{
	\begin{aligned}
		K (h_{\text{open}} - h)  & \quad \hat{d} \geq \hat{d}_c \\
		K (h_{\text{closed}} - h) & \quad \hat{d} < \hat{d}_c    
	\end{aligned}
	\right.
\end{equation}

\noindent where $h$ is the hand closure, $\hat{d}$ is the normalized relative hand-object distance, $K$ is the controller gain, $\hat{d}_c$ is the critical normalized relative hand-object distance, and $h_{\text{open}}$ and $h_{\text{closed}}$ are the reference values for the controller in the first and second phases, respectively. For the simulated-trajectory environment, there existed a single tuple of the policy parameters $(K,\hat{d}_c)$ which resulted in the maximum reward for almost all samples of the environment parameters. However, in DEXTRON, a single controller parameter tuple results in successful grasps in a few samples of the environment.

We define the set of the above expert policies as:

\begin{equation}
        \begin{split}
                \Pi_E = \Big\{ & \pi_e(h, \hat{d}) \big| \\
                               & K \in [K_{min}, K_{max}] \: \& \: \hat{d}_c \in [\hat{d}_{c,min}, \hat{d}_{c,max}] \Big\}
        \end{split}
\end{equation}

\noindent where $K_{min}$, $K_{max}$, $\hat{d}_{c,min}$, and $\hat{d}_{c,max}$ are constants given in Table~\ref{tab:param_dists}. We run a Monte Carlo simulation to find a working expert policy $\pi_e \sim \mathcal{U}(\Pi_E)$ for every sample from the environment. For this, both the environment settings and the policy parameters are sampled each time, the policy is rolled out in the simulation environment. If the sum of rewards for that rollout is greater than $1$, we store the environment and the policy sampled parameters tuple, $\Big( (\lambda, \delta_x, \delta_y, t_n), (K, \hat{d}_c) \Big)$, in $G_s$, a database of \textit{successful} policy-environment parameters. It is worth noting that for some environment settings, there exist no policies $\pi_e \in \Pi_E$ that result in successful grasps. This is due to two possible reasons. First, $\Pi_E$ class of policies may not express a policy that works for some environment settings, i.e. the robot maneuvers required for timely grasping of the object without collision is not represented in $\Pi_E$. Second, some reaching trajectories collide with the object unconditionally, no matter what policy is being used, due to relative hand/object pose.

\begin{table}[h!]
	\centering
	\caption{Controller and environment parameters}
	\label{tab:param_dists}
	\begin{tabular}{lc} 
		\toprule
		\textbf{Parameter} & \textbf{Value} \\ [0.5ex]
		\midrule
                \textbf{Environment} & \\
                Offset X ($\Delta_x$) & $[-0.06,0.06]$ \\
                Offset Y ($\Delta_y$) & $[-0.06,0.06]$ \\
                Time scale ($t_s$) & 2.5 \\
                Time noise mean ($\mu_t$) & 0.5\\
                Time noise variance ($\sigma_t^2$) & $0.8^2$ \\ [1.5ex] % T = t * t_s + t_n
                \textbf{Controller} & \\
                Controller gain $\big[K_{min},K_{max}\big]$ & $[1,1.5]$ \\
                Controller threshold $\big[\hat{d}_{c,min},\hat{d}_{c,max}\big]$ & $[0.01,0.90]$ \\ 
                Hand open target ($h_{\text{open}}$) & 0 \\
                Hand closed target ($h_{\text{closed}}$) & 0.8 \\ [1ex]

		\bottomrule
	\end{tabular}
\end{table}

%%%%%%%%%%%%%%%%%%%%%%%%%%%%%%%%%%%%%%%%%%%%%%%
\subsection{Learning End-to-end Policy \label{sub:learning_end_to_end_policy}}

Here we provide our main method for training the end-to-end environment and some baselines.

\subsubsection{RLIL}
Like our previous study \cite{Sharif2020-ll}, we use Soft-Actor Critic (SAC) \cite{Haarnoja2018-ts} method for the RL. As detailed in \cite{Sharif2020-ll}, our problem is an instance of a Keylock MDP with hard-to-explore sparse and delayed rewards. SAC is a maximum entropy reinforcement learning method which offers good exploration for finding the optimal solution. However, using RL alone may result in very high variance and sensitivity to random seeds. Accordingly, we use a method to guide the explorations, hence our use of IL besides RL. Using IL, RL finds rewarding states which otherwise could never be found due to the large state space dimensions. Since SAC is an off-policy method and off-policy methods do not care about the policy used to generate the transitions in a replay buffer, we can simply store the expert policy transitions in the same buffer as the agent. For the details of the method one may refer to \cite{Sharif2020-ll}. Also, the architecture of value, action-value, and policy networks are the same as \cite{Sharif2020-ll}. The hyper-parameters are given in Table~\ref{tab:hyperparams}. We provide the results of a no-IL case (pure RL) along with other results as a baseline. We call the policy trained using RLIL or RL alone $a = \pi_{RL} (s)$ for future reference.

\subsubsection{Behavior Cloning}
As a baseline, we provide the results of Behavior Cloning (BC) which is training a policy $a=\pi_{BC}(s)$ in a supervised manner on a dataset including transition tuples $(s,a)$ from rolling out the expert policies and environment settings sampled from $G_s$. The policy $\pi_(BC)$ is represented as a Neural Network having the same structure as the policy network of the RL. A scheduled learning rate with Adam optimizer was used for training.

\subsubsection{Testing Simulated-Trajectory Environment Policy in DEXTRON}
As a comparison, we also trained a policy in the simulated-trajectory environment, and tested it in DEXTRON. The policy was trained using the same settings as described in \cite{Sharif2020-ll}.

\begin{table}[h!]
	\centering
	\caption{Hyperparameter selection}
	\label{tab:hyperparams}
	\begin{tabular}{lc} 
		\toprule
		\textbf{Hyperparameter} & \textbf{Value} \\ [0.5ex]
		\midrule
		\textbf{SAC} & \\
		    Learning rate & 3e-4 \\
            Agent replay buffer size & 1e6 \\
            Demo replay buffer size & 1e6 \\
		    Mini-batch size & 32 \\
		    Polyak coefficient & 0.01 \\
		    Epoch size & 1000 frames \\
            Discount factor ($\gamma$) & 0.99 \\
            Warm start & 10000 frames \\
            Network hidden size & 256 \\[1.5ex]
                
		\textbf{Success Model \& Behavior Cloning} & \\
            Learning rate & 1e-3 \\
            Learning rate decay & 0.5 every 100 epochs \\
            Optimizer & Adam \\
            Batch size (Success Model) & 1024 \\
            Batch size (Behavior Cloning) & 512 \\ [1.5ex]

		\bottomrule
	\end{tabular}
\end{table}

%%%%%%%%%%%%%%%%%%%%%%%%%%%%%%%%%%%%%%%%%%%%%%%
\subsection{Success Model \label{sub:success model}}

In this section, we propose a model that can add to the transparency of the trained RL policy, $\pi_{RL}$, by evaluating its selected actions based on actions which are normally selected by the expert policies in $G_s$ in similar similar states. As mentioned in previous sections, we collected a database, $G_s$, of specialized Markovian policies for each environment instance. Existence of a specialized \textit{Markovian} policy for each sample of a stochastic Environment does not guarantee the existence of a generic Markovian policy that performs as good as the specialized policies, which makes reasoning about the baseline performance of this problem hard. This can happen, for instance, due to partial observability in our stochastic environment with our current system states. Based on this fact, it may not be fair to compare the performance of our policy obtained by RLIL with the specialized expert policies performance, but we can still expect to see valuable information by comparing them collectively with our policy at each timestep. For this purpose, we propose a predictive network $\Omega(s,a)$ which is essentially a classifier that predicts the chance of final terminal success in state $s$ if we take action $a$. By success, we mean obtaining a sum of rewards of $20$ in the episode, i.e. $\sum{r}=20$. This model is inspired by Levine et al. \cite{Levine2016-cq} in which they proposed a predictive model to take actions at each timestep that are expected to give the best final score for grasping.

We train the success model, $\Omega(s,a)$, in a supervised manner. We create a dataset, $\Upsilon$, of tuples, $(s,a,o)$, which include the state, action, and final outcome of rolling out a policy in the environment. The final outcome, $o$, is defined as $1$ if we get a sum of rewards of $20$ and $0$ otherwise. The dataset, $\Upsilon$, is a union of two smaller datasets, $\Upsilon_{EX}$ and $\Upsilon_{RL}$, where $\Upsilon_{EX}$ includes transitions from rolling out sampled policy $\pi_e \sim \mathcal{U}(\Pi_E)$ in a sampled environment, and $\Upsilon_{RL}$ includes transitions from trained RL policy roll-outs in an environment setting sampled from $G_s$. The $\Upsilon_{EX}$ dataset would normally be highly imbalanced leaning towards failure cases, with a ratio of failure cases to success cases of about $50:1$. We use oversampling technique to overcome this imbalance, by sampling the policy (and the environment settings) half-of-the-times from $G_s$, and half-of-the-times from $\Pi_E$ (and the general set of environment parameters). This ensures a balance between success and failure cases in the dataset. The purpose of including $\Upsilon_{RL}$ in the training dataset is to improve the predictive feature of $\Omega(s,a)$ over $\pi_{RL}$. This ensures $\Omega(s,a)$ can explain the behavior of the RL agent with respect to the collective expert policies while providing better predictions on the RL policy itself.

Here, $\Omega(s,a)$ is implemented as a multi-layer perceptron (MLP). We use 3 layers of a fully connected network with ReLU activation functions. The last layer output is of dimension 2 for two classes, success and failure, which can be later fed into a SoftMax for normalization. Cross Entropy loss function is used to calculate the difference between labels and network predictions. The network is then trained using back-propagation and Adams optimization method \cite{Kingma2014-fb}. During training, $20\%$ of the data points were separated as test data and $80\%$ were set aside for training.

%%%%%%%%%%%%%%%%%%%%%%%%%
%%     EXPERIMENTS     %%
%%%%%%%%%%%%%%%%%%%%%%%%%
\section{Experiments and Simulations}
\label{sec:experiments_and_simulations}
%%%%%%%%%%%%%%%%%%%%%%%%%%%%%%%%%%%%%%%%%%%%%%%
\subsection{Experimental Setup \label{sub:experimental_setup}}

\begin{figure}[tpb]
        \centering
        \includegraphics[width=0.45\textwidth]{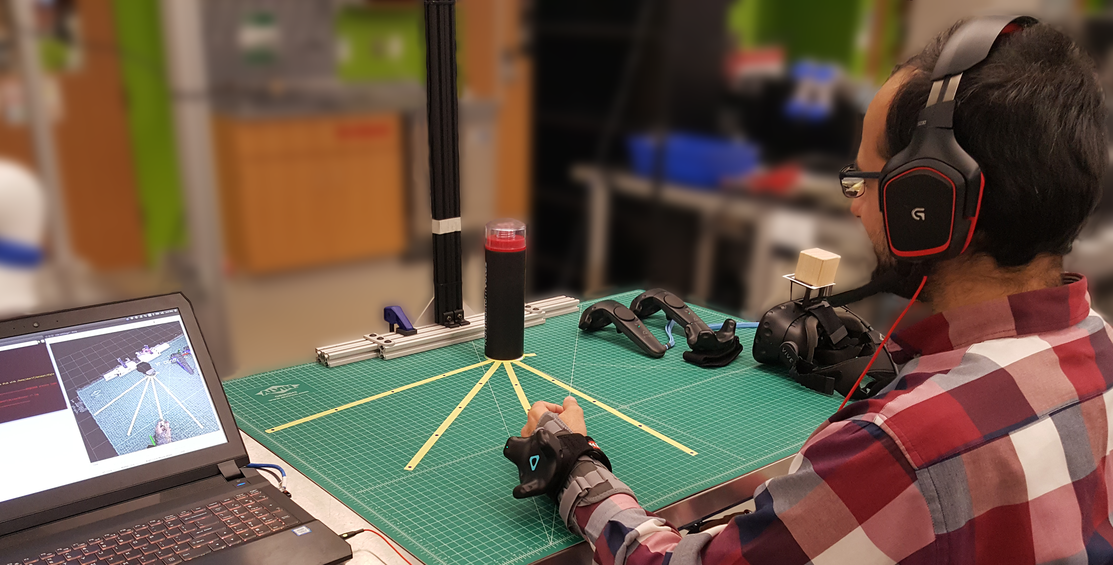}
        \caption{The experiment setup with a subject.}
        \label{fig:subject}
\end{figure}

In order to collect real human hand trajectory data, we set up an experiment with a top-mount Primesense Carmine 1.09 RGB-D camera, an HTC Vive Tracker (2018) with HTC Vive Base Station 1.0\footnote{https://www.vive.com/}, a headphone, and a wrist brace (Fig.~\ref{fig:workspace}). The HTC Vive Tracker is reported to have sub-millimeter precision and an average accuracy of about 2 mm in normal working conditions \cite{Borges2018-aa}. No further filtering was done in addition to the builtin proprietary processing of the tracking data. Subjects wore the headphone, the wrist brace, and an HTC Vive Tracker on their left hand (Fig.~\ref{fig:subject}). The HTC Vive Headset was left on the workspace during the experiment. The workspace was marked with 16 markers evenly distributed on a $4 \times 4$ polar grid centered at the object's center in order to specify the start position of the reaching trajectories. Each subject had to perform 4 trials where each trial consisted of 16 reaching actions: start reaching from a marker after hearing a beep sound, grasp the object, lift the object up to a specified threshold, place the object where it was, retract the hand to the position of the next marker, and wait for the next beep sound. The HTC Vive Tracker recorded the pose of the left hand of the subject. The RGB-D camera recorded the point cloud and the RGB image of the experiment. The coordinate frames of the HTC Vive Tracker and the RGB-D camera were registered on to each other by using an April Tag \cite{Olson2011-zk} attached to the HTC Vive Headset as a connection intermediate frame known in both coordinate frames. ROS was used as the back-end system. Data were recorded in ROS's Bag file format.

%%%%%%%%%%%%%%%%%%%%%%%%%%%%%%%%%%%%%%%%%%%%%%%
\subsection{Data Collection and Processing \label{sub:data_collection_and_processing}}

In total, trajectory data from 9 male subjects were recorded. Due to different hand sizes of the subjects, each trajectory file was visually inspected and manually offset so the pose of the human's hand in the depth sensor coordinate frame approximately matched the pose of the simulated robot hand model based on the Tracker's information. Exact matching was not pursued since this was compensated later in the Monte Carlo simulation by randomizing the trajectory offsets, $\delta_x$ and $\delta_y$. Each Bag file was annotated manually for the important time instances, onset of the reaching trajectory, $\theta_1$, the instance of hand actually reaching the object just before the grasp, $\theta_2$, object reaching its maximum height in the lifting phase, $\theta_3$, the releasing moment, $\theta_4$, and hand arriving at the rest position again, $\theta_5$. An automated script then segmented each bag file into individual reaching trajectories. Each extracted trajectory included from $\theta_1$ to $\theta_3$ instances. Totally, $9 \times 16 \times 4 = 576$ individual trajectories were obtained. The whole data collection was before the COVID-19 pandemic hence no health protocols were required.

%%%%%%%%%%%%%%%%%%%%%%%%%%%%%%%%%%%%%%%%%%%%%%%
\subsection{Simulations \label{sec:simulations}}

We parallelized the execution of the Monte Carlo simulation using 6 nodes on a computer cluster, with each node running 25 simultaneous threads at the same time. This took about 12 hours for each node to test 3,068,600 episodes in total from which 71798 episodes resulted in a sum of reward greater than 1 (2.3\% rate of success).

We used Digideep \cite{Sharif2019-ee} for its implementation of the SAC method. We utilized PyTorch \cite{Paszke2019-ty} for the Neural Networks and MuJoCo \cite{Todorov2012-ty} for physics modeling. The environment of MuJoCo HAPTIX \cite{Kumar2015-cm} was used as a base for our simulation environments. DeepMind's dm\_control was adopted as a Python wrapper for MuJoCo. The model used for the robot hand was built based on the Ottobock BeBionic \cite{Medynski2011-qg} robot prosthetic hand in Solidworks, then exported to MuJoCo. Training our RL models took about 25 hours on a single NVIDIA P100 GPU. Generating the datasets for training the success model took between 2 to 6 hours using 20 threads simultaneously. The $\Upsilon$ dataset included ~30M transitions which were from running 180K episodes and it required ~18 GB for storage. The ratio of success labels to total number of transitions was $54\%$.  The dataset for training the BC policy included ~10M transitions from running ~45K episodes.

%%%%%%%%%%%%%%%%%%%%%%%%%
%%       RESULTS       %%
%%%%%%%%%%%%%%%%%%%%%%%%%
\section{Results}
\label{sec:results}

\begin{figure}[h]
	\centering
	\begin{tabular}[c]{cc}
		\begin{subfigure}{0.21\textwidth}
			\centering
			\includegraphics[width=\linewidth]{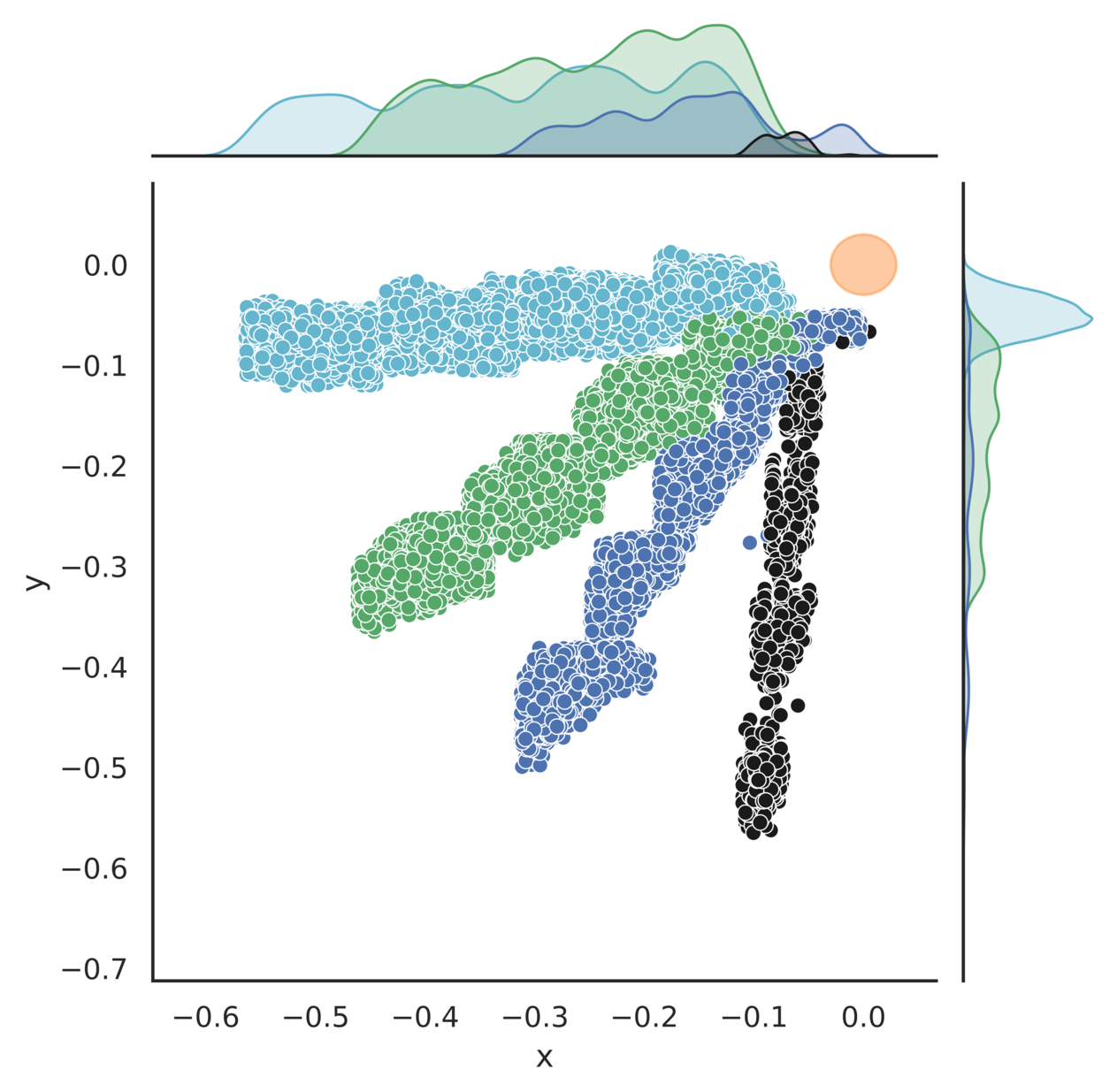}
			\caption{Start locations after MC.}
			\label{fig:trajectories_scatter}
		\end{subfigure}&
		\begin{subfigure}{0.26\textwidth}
			\centering
			\includegraphics[width=\linewidth]{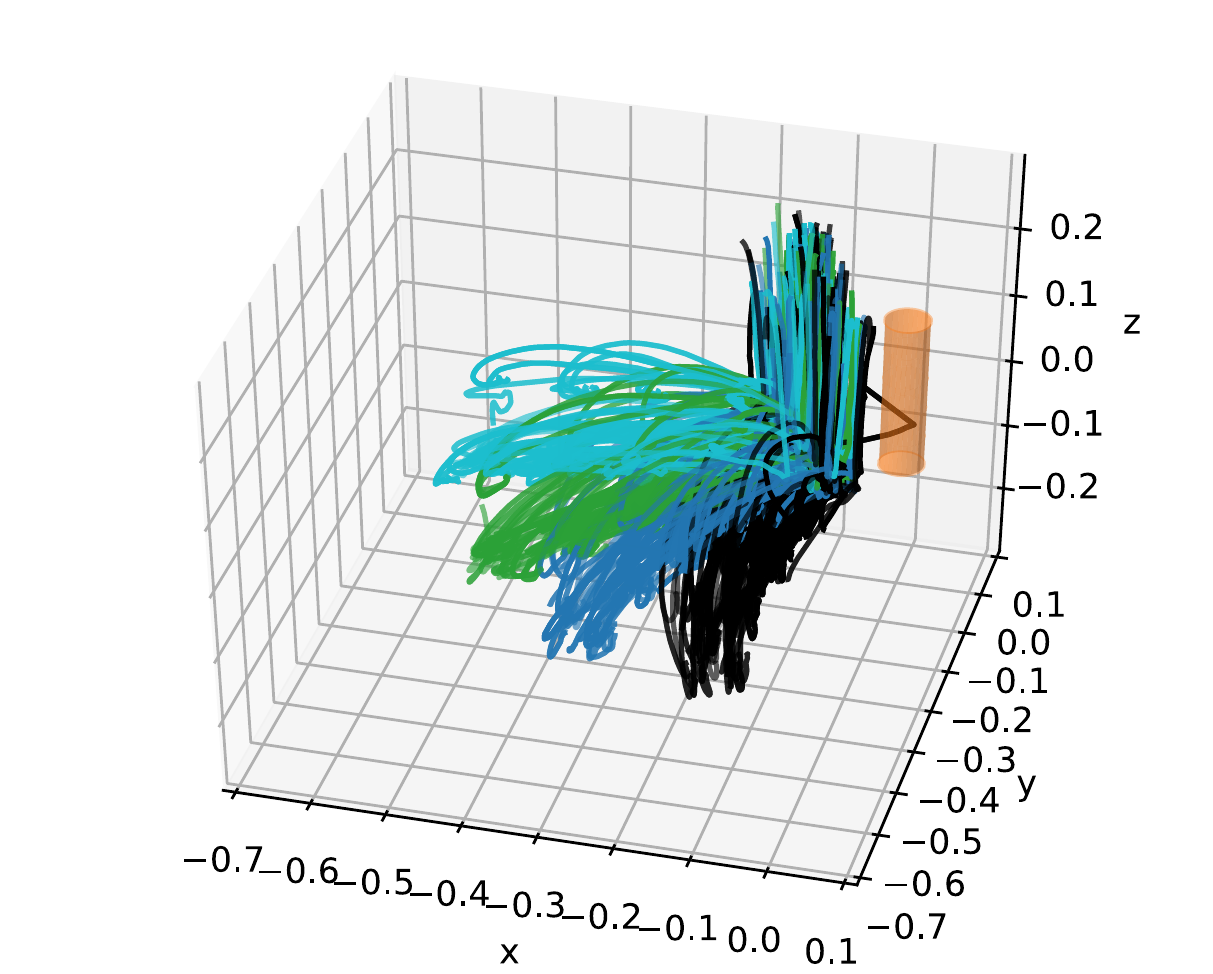}
			\caption{Original trajectories before MC.}
			\label{fig:trajectories_3d}
		\end{subfigure}\\
	\end{tabular}
	
	\caption{Sample successful trajectories of the MC simulation. Object is located at $x=0$ and $y=0$.}
	\label{fig:trajectories}
\end{figure}

By using the Monte Carlo simulation, we could obtain the distribution over the randomized controller and environment parameters. Fig.~\ref{fig:trajectories} shows the distribution of the trajectories in 3d (before MC) and starting points of successful trajectories (after MC). It is worth noting that the region indicated with black trajectories and dots included the least successful grasps, probably due to the narrow angle of hand approach with respect to the object, which results in a collision most of the time.

\begin{figure}[bht]
    \centering
    \begin{tabular}[t]{cc}
        \begin{subfigure}{0.33\textwidth}
            \centering
            \smallskip
            \includegraphics[width=0.99\linewidth]{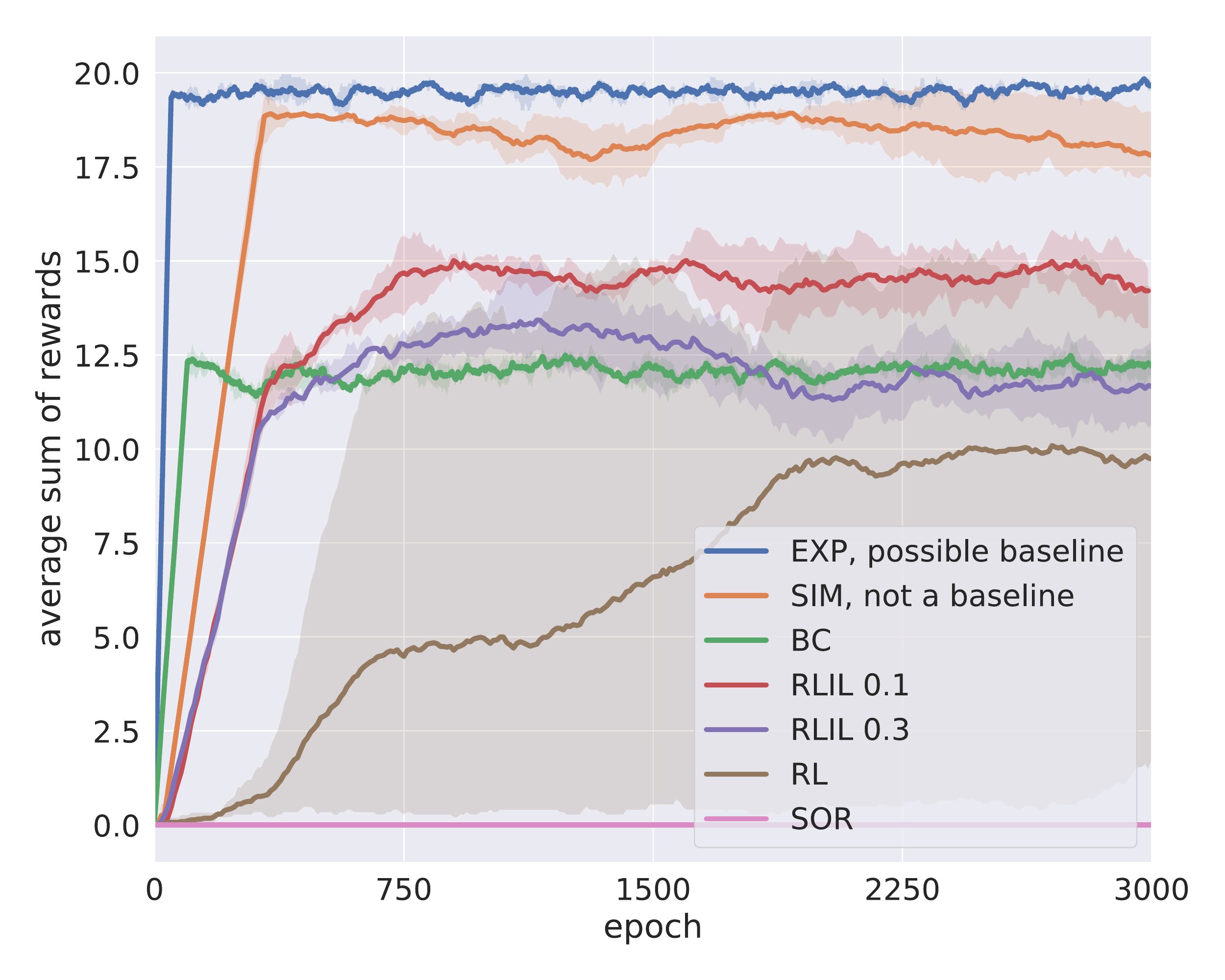}
			\caption{Baseline comparison} %{Light Unit}
			\vspace{1\baselineskip}
            \label{fig:baselines}
        \end{subfigure}
        &
        \begin{tabular}{c}% if you add [t], than sub images are pushed down
            % \smallskip
			\hspace{-2.5em}
            \begin{subfigure}[t]{0.15\textwidth}
                \centering
                % \vspace{-0.5\baselineskip}
				\includegraphics[width=0.99\textwidth]{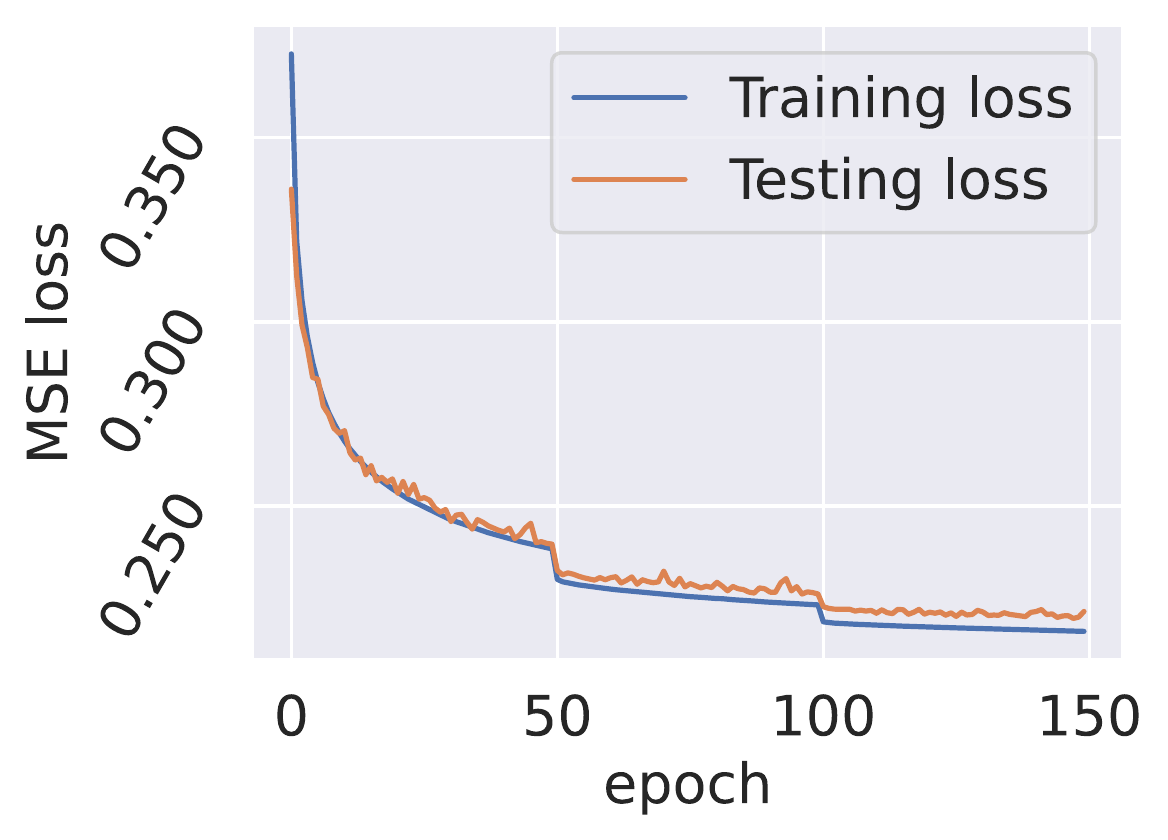}
				\caption{Success model}
				\vspace{0.5\baselineskip}
                \label{fig:learning_curve_sm}
            \end{subfigure}\\
            \hspace{-2.5em}
            \begin{subfigure}[t]{0.15\textwidth}
                \centering
				% \vspace{-1.4\baselineskip}
                \includegraphics[width=0.99\textwidth]{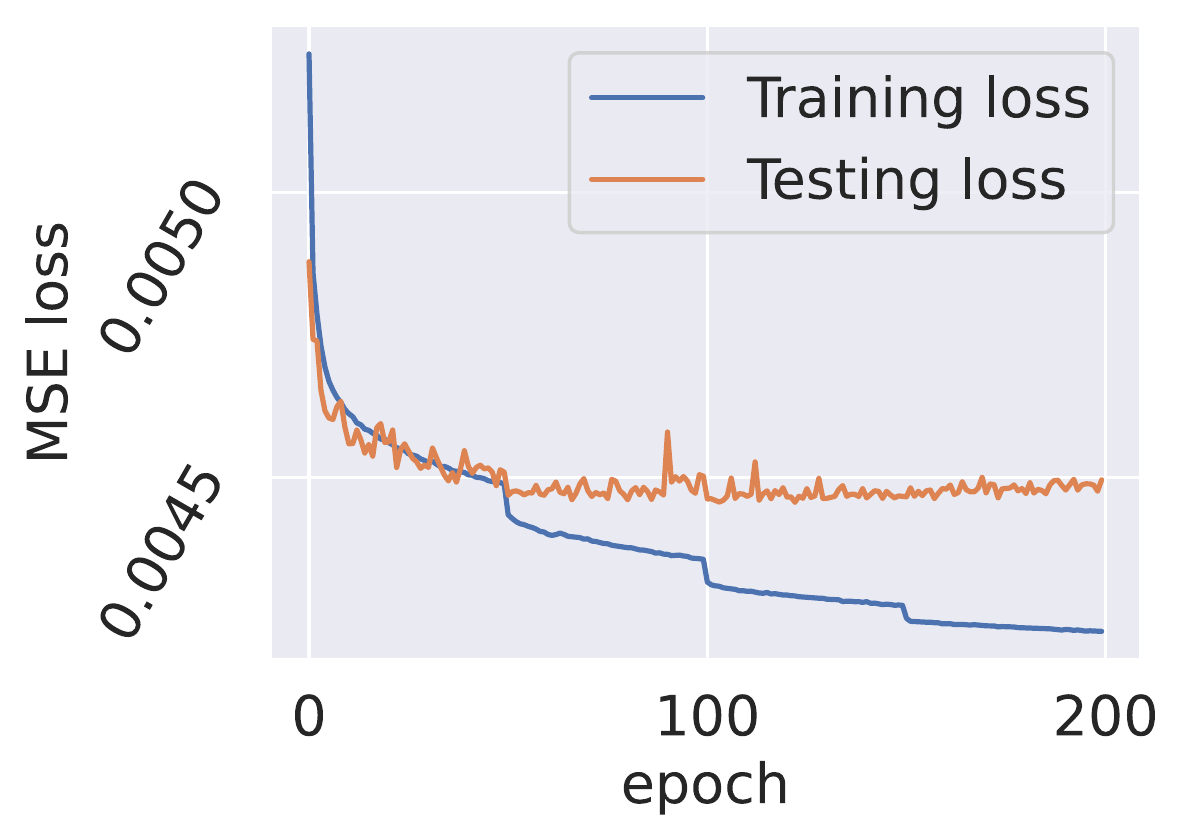}
				\caption{Behavior cloning}
				\vspace{0.5\baselineskip}
                \label{fig:learning_curve_bc}
            \end{subfigure}
        \end{tabular}\\
    \end{tabular}
    \caption{Learning curves}
    \label{fig:learning_curves}
\end{figure}

% Baselines and results.
The sum of rewards for several cases are demonstrated in Fig.~\ref{fig:baselines}. The expert policies (stored in $G_s$) are shown as ``EXP'' in the plot. Note that although only $r=20$ cases were included in the $G_s$ database, the result of replaying those policies in the environment sometimes did not give us a sum of reward of $20$. We could not find the source for this discrepancy, but we guess it might be related to some stochasticity in the MuJoCo environment which causes slightly different results based on different environment random seeds. The reproduced results of \cite{Sharif2020-ll} using the new reward function are shown as ``SIM''. This is obviously not a baseline since it is trained on a different environment (simulated-trajectory environment). The BC sum of reward is shown under ``BC''. Also, the learning curve of the BC method is shown in Fig.~\ref{fig:learning_curve_bc}. The effect of Demo-Use Ratio (DUR), i.e. the ratio of demo transitions sampled for training as defined in \cite{Sharif2020-ll}, is shown by ``RLIL 0.1'' for $DUR=0.1$ and ``RLIL 0.3'' for $DUR=0.3$. The pure RL case is shown as ``RL''. Rolling out the policy of ``SIM'' in DEXTRON is shown as ``SOR''. The results are shown for 3 different random seeds. Every epoch includes 1000 frames.

\begin{figure*}[t]
	\centering
	\begin{tabular}[c]{ccccc}
		\begin{subfigure}{0.18\textwidth}
			\centering
			\includegraphics[width=\linewidth]{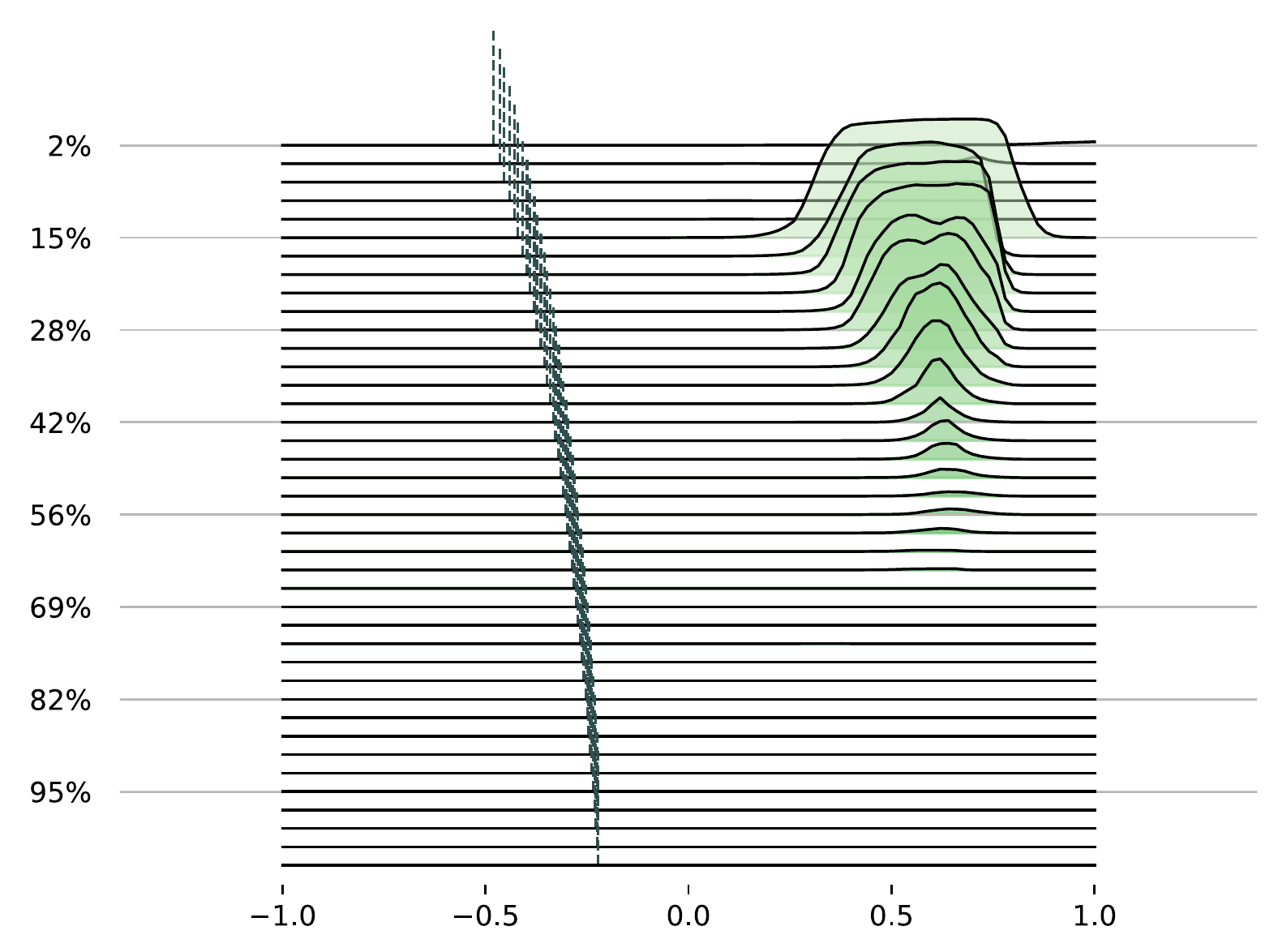}
			\caption{Disappearing hope window due to continued wrong choice.}
			\label{fig:joy_a_rand_disappearing}
		\end{subfigure}&
		\begin{subfigure}{0.18\textwidth}
			\centering
			\includegraphics[width=\linewidth]{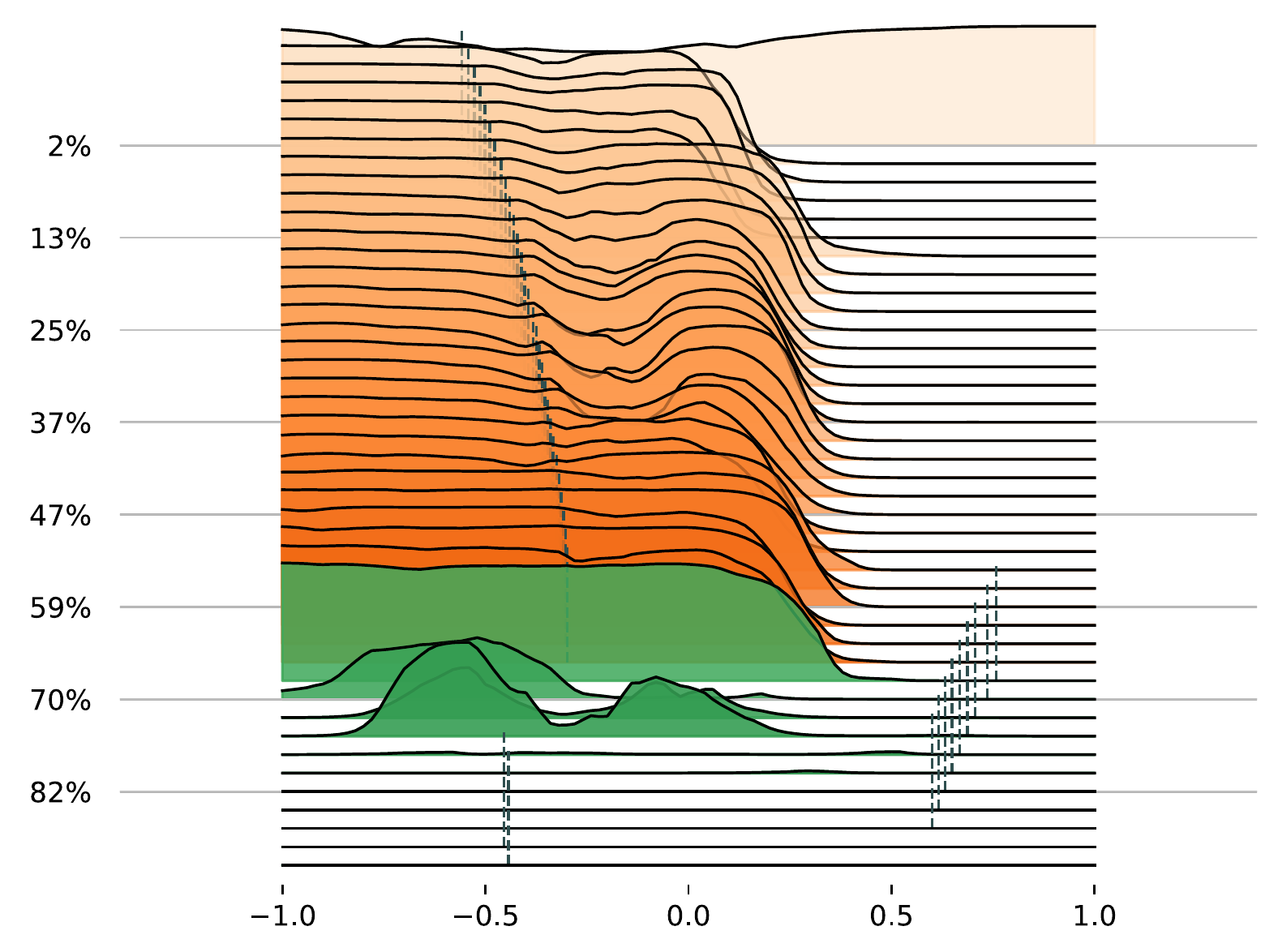}
			\caption{Early closing command caused failure.}
			\label{fig:joy_b_rand_early}
		\end{subfigure}&
		\begin{subfigure}{0.18\textwidth}
			\centering
			\includegraphics[width=\linewidth]{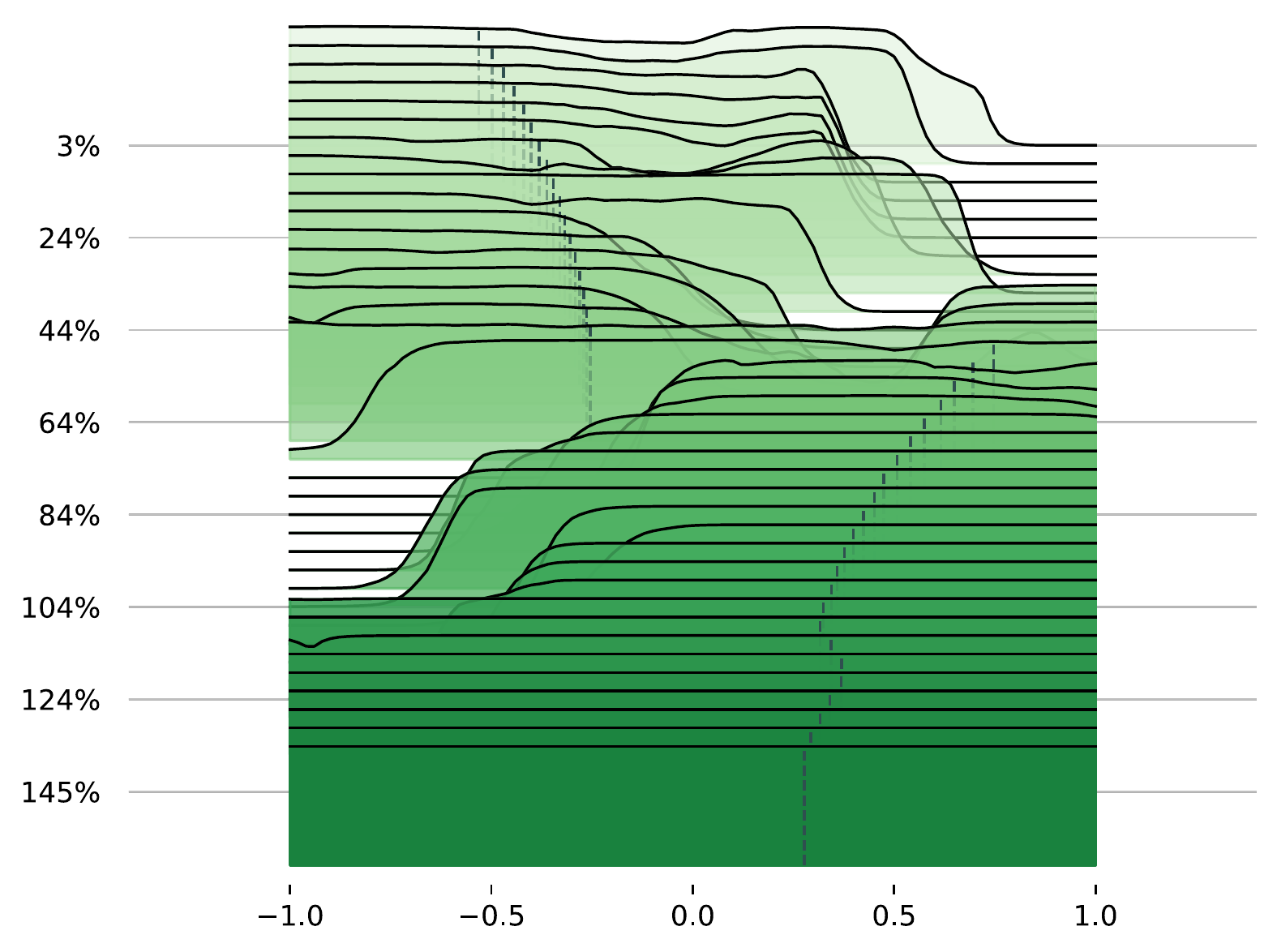}
			\caption{Correct prediction for expert policy.}
			\label{fig:joy_c_demo_fine}
		\end{subfigure}&
		\begin{subfigure}{0.18\textwidth}
			\centering
			\includegraphics[width=\linewidth]{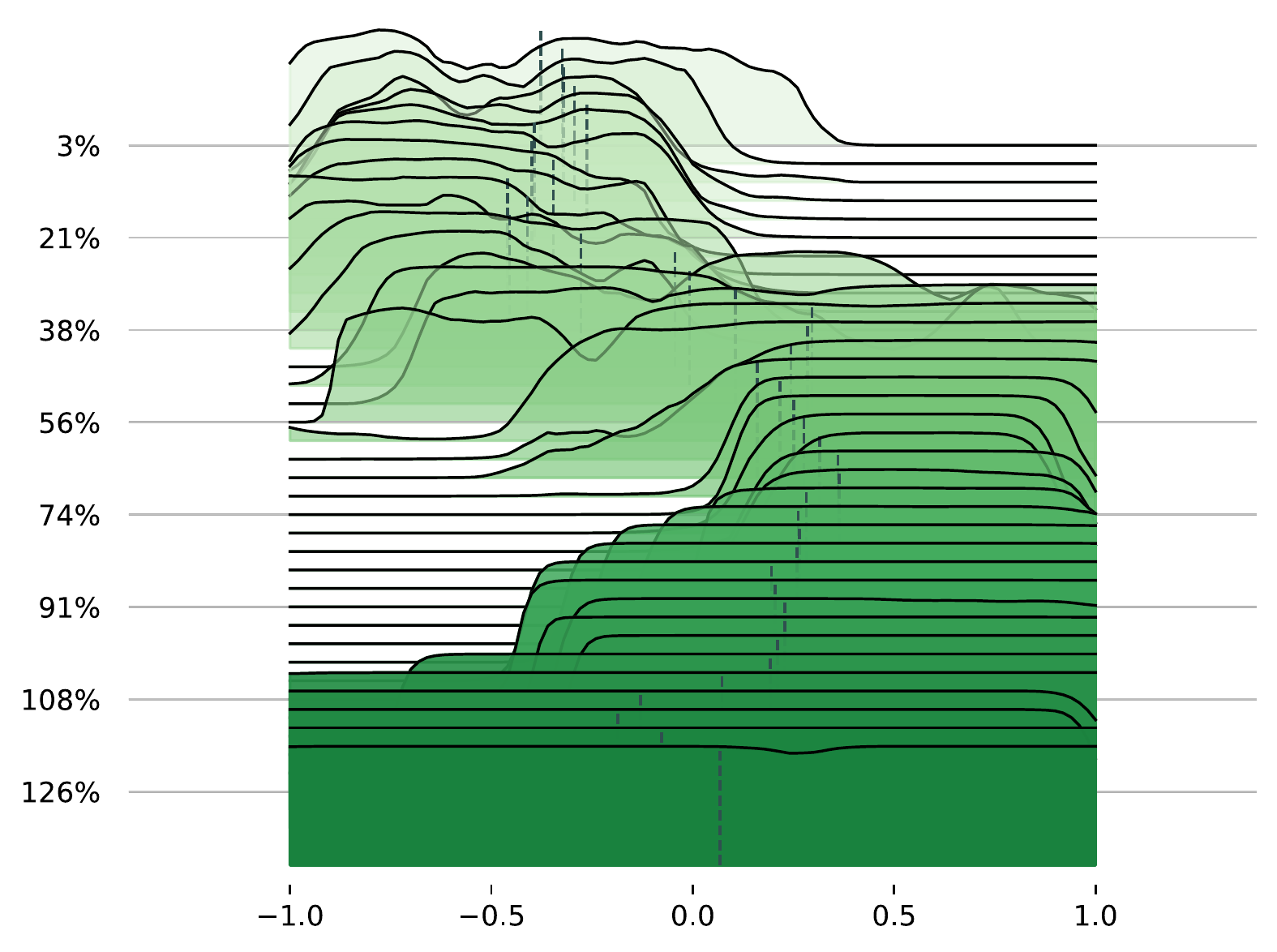}
			\caption{Correct prediction for RL policy.}
			\label{fig:joy_d_rl_fine}
		\end{subfigure}&
		\begin{subfigure}{0.18\textwidth}
			\centering
			\includegraphics[width=\linewidth]{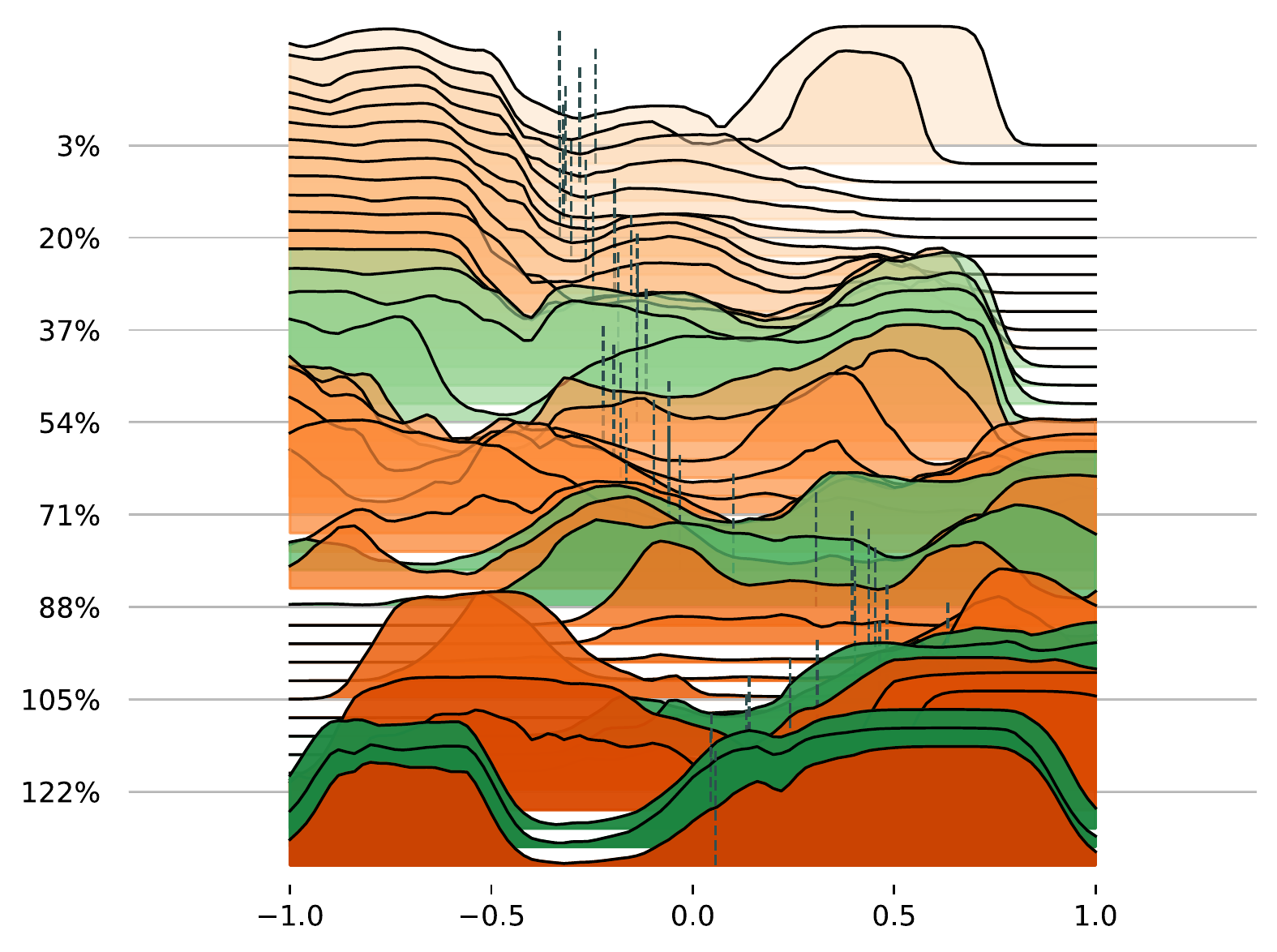}
			\caption{A case in which the prediction model fails.}
			\label{fig:joy_e_rl_bad}
		\end{subfigure}\\
	\end{tabular}
	
	\caption{Applying the success model on five sample trajectories. The vertical axis shows the normalized timestep of the trajectory with respect to its time-of-reach. The horizontal axis shows the action. The short lines show the action that was actually chosen by the policy. The \fcolorbox{black}{green}{\rule{0pt}{4pt}\rule{4pt}{0pt}} color shows a true prediction of the final outcome. The \fcolorbox{black}{red}{\rule{0pt}{4pt}\rule{4pt}{0pt}} color shows misclassifications.}
	\label{fig:joyplots}
\end{figure*}

The success model inference plots are shown in Fig.~\ref{fig:joyplots}. At each state $s$, we calculate the value of the function for a discretized set of actions $a\in \{-1,-0.96,...,0.96,1\}$, and draw the success curve for that time instance. The graph shows the rate of final success we expect in state $s$ if we perform action $a$. Five cases are displayed in Fig.~\ref{fig:joyplots}.

%%%%%%%%%%%%%%%%%%%%%%%%%
%%     DISCUSSION      %%
%%%%%%%%%%%%%%%%%%%%%%%%%
\section{Discussion}
\label{sec:discussion}

In this work, we offered training an end-to-end policy for human-in-the-loop robot grasping via RL. In a previous work \cite{Sharif2020-ll}, we provided an end-to-end policy for a simulated-trajectory environment, while in this work a real-trajectory environment (i.e. DEXTRON) was considered. From the ``SOR'' plot in Fig.\ref{fig:baselines}, it is noticed that the policy is sensitive to the trajectories it sees during training: the policy trained in the simulated-trajectory environment does not transfer to DEXTRON successfully.

``RLIL 0.1'' provides the best result of all other methods plotted in Fig.\ref{fig:baselines}, where it has the highest average sum of rewards and reasonable variance (w.r.t. ``RL'' method). The ``BC'' performed moderately as its policy is only trained on expert data and has no chance of interaction with the environment. Contrary to \cite{Sharif2020-ll} where pure RL could not learn anything, ``RL'' learns useful policies in DEXTRON, probably due to stochasticity and higher variance in the environment. Here, we could reach a maximum score of $15/20$ from among all of our different tested methods, i.e. a success rate of $75\%$ in grasping the objects. This is certainly not desirable as an end product; however, it provides a good starting point and baseline for future extensions of this work.

The success model provides a post-hoc explanation about the operation of the RL policy network. This is significant since it provides a way to compare the quality of the RL policy with the set of expert policies collectively. The results of applying the success models on 5 sampled policy roll-outs is demonstrated in Fig.~\ref{fig:joyplots}. The success model has successfully predicted a failure outcome in Fig.~\ref{fig:joy_a_rand_disappearing} in all timesteps; where a success window (the bumps in the graph) existed in the early stages of the trajectory, but that diminishes as the action is constantly chosen far from that window. In Fig.\ref{fig:joy_b_rand_early}, the final failure outcome is not attributed to the policy's actions in the early stages of the trajectory, rather, it is because of a too early closing command at about ~70\% of the trajectory which results in a collision with the object and diminishes all chances of recovery quite fast. In Fig.~\ref{fig:joy_c_demo_fine} and Fig.~\ref{fig:joy_d_rl_fine} the success model results are visualized for two successful outcomes from the expert and the trained RL policy (from ``RLIL 0.1''). We see how appropriate choices of actions by policies' have resulted in a successful final outcome. It is important to note though, like any model, the success model can fail to provide a useful explanation (Fig.~\ref{fig:joy_e_rl_bad}).

By providing predictions into the future, the success model may also be used in real time to give feedback to the user about the outcomes of being at a specific state. While the success model acts like an action-value function, it is still different in that it is trained on failure cases as well. So, the success model can be thought to provide failure-/ success-awareness for the policy. 

As our best method reaches a maximum score in about 750 epochs (Fig.\ref{fig:baselines}), which means 750k frames or ~4 hours of experiment, it is feasible to learn the policy directly in real life. One may also use sim2real methods \cite{Sadeghi2016-xt, Tobin2017-ew} to transfer the policy to real life. The current dataset and environment, considers only the reaching part of the trajectory which limits its use in real life unless we create mechanisms to detect onset of the reaching action. Also, the current approach does not include any releasing. In future works, we plan to extend the current approach to learning end-to-end policies for more complex tasks which may contain several grasps/releases. Including a variety of objects, learning from images to relax the need for raw system states, and learning shared control strategies to use EMG signals along with the system states are a few other suggestions for future works. The last improvement can relax our assumption about knowing human intent \textit{a priori} by providing a shared control framework.

%%%%%%%%%%%%%%%%%%%%%%%%%
%%     CONCLUSION      %%
%%%%%%%%%%%%%%%%%%%%%%%%%
\section{Conclusion}
\label{sec:conclusion}

In this work we offer the methodology to train end-to-end policies for human-in-the-loop robot grasping on real reaching trajectories. The fact that human moves the robot makes this problem naturally fall into the Human-Robot Collaboration category. We trained our policy using real human trajectories which were augmented using Monte Carlo simulation. The fact that our method does not rely on EMG makes it suitable for lengthy and repetitive tasks; however, the fact that it has pretty small information about the human intent makes it less readily available for generic applications. We hope that a hybrid method integrating our method with EMG/ENG inference can take advantage of both methods while reducing robustness issues of the EMG-based control methods. Also, by providing DEXTRON, we hope to attract the attention of the RL and HRC communities to the human-in-the-loop robot grasping problem.


\section*{ACKNOWLEDGMENT}
We would like to thank Eoin Daly for his help with data processing for the real trajectories.

%%%%%%%%%%%%%%%%%%%%%%%%%%%%%%%%%%%%%%%%%%%%%%%%%%%%%%%%%%%%%%%%%%

% \addtolength{\textheight}{-12cm}  

\bibliographystyle{hieeetr}
% \bibliography{IEEEabrv,references}
\bibliography{ms}

\begin{thebibliography}{10}

\bibitem{Muceli2014-lx}
S.~Muceli, N.~Jiang, and D.~Farina, ``Extracting signals robust to electrode
  number and shift for online simultaneous and proportional myoelectric control
  by factorization algorithms,'' {\em IEEE Trans. Neural Syst. Rehabil. Eng.},
  vol.~22, pp.~623--633, May 2014.

\bibitem{Hakonen2015-rt}
M.~Hakonen, H.~Piitulainen, and A.~Visala, ``Current state of digital signal
  processing in myoelectric interfaces and related applications,'' {\em Biomed.
  Signal Process. Control}, vol.~18, pp.~334--359, Apr. 2015.

\bibitem{Hwang2017-to}
H.-J. Hwang, J.~M. Hahne, and K.-R. M{\"u}ller, ``Real-time robustness
  evaluation of regression based myoelectric control against arm position
  change and donning/doffing,'' {\em PLoS One}, vol.~12, p.~e0186318, Nov.
  2017.

\bibitem{Chadwell2016-zk}
A.~Chadwell, L.~Kenney, S.~Thies, A.~Galpin, and J.~Head, ``The reality of
  myoelectric prostheses: Understanding what makes these devices difficult for
  some users to control,'' {\em Front. Neurorobot.}, vol.~10, p.~7, Aug. 2016.

\bibitem{Dosen2010-tg}
S.~Do{\v s}en, C.~Cipriani, M.~Kosti{\'c}, M.~Controzzi, M.~C. Carrozza, and
  D.~B. Popovi{\'c}, ``Cognitive vision system for control of dexterous
  prosthetic hands: Experimental evaluation,'' {\em J. Neuroeng. Rehabil.},
  vol.~7, p.~42, Aug. 2010.

\bibitem{Gigli2017-ih}
A.~Gigli, A.~Gijsberts, V.~Gregori, M.~Cognolato, M.~Atzori, and B.~Caputo,
  ``Visual cues to improve myoelectric control of upper limb prostheses,'' Aug.
  2017, 1709.02236.

\bibitem{Sharif2019-sk}
M.~Sharif, D.~Erdogmus, and T.~Padir, ``Particle filters vs hidden markov
  models for prosthetic robot hand grasp selection,'' {\em International
  Journal of Robotic Computing}, vol.~1, p.~25, July 2019.

\bibitem{Zhuang2019-td}
K.~Z. Zhuang, N.~Sommer, V.~Mendez, S.~Aryan, E.~Formento, E.~D'Anna,
  F.~Artoni, F.~Petrini, G.~Granata, G.~Cannaviello, W.~Raffoul, A.~Billard,
  and S.~Micera, ``Shared human--robot proportional control of a dexterous
  myoelectric prosthesis,'' {\em Nature Machine Intelligence}, vol.~1,
  pp.~400--411, Sept. 2019.

\bibitem{Santello1998-xb}
M.~Santello and J.~F. Soechting, ``Gradual molding of the hand to object
  contours,'' {\em J. Neurophysiol.}, vol.~79, pp.~1307--1320, Mar. 1998.

\bibitem{Smith2011-fz}
L.~H. Smith, L.~J. Hargrove, B.~A. Lock, and T.~A. Kuiken, ``Determining the
  optimal window length for pattern recognition-based myoelectric control:
  balancing the competing effects of classification error and controller
  delay,'' {\em IEEE Trans. Neural Syst. Rehabil. Eng.}, vol.~19, pp.~186--192,
  Apr. 2011.

\bibitem{Bouwsema2010-me}
H.~Bouwsema, C.~K. van~der Sluis, and R.~M. Bongers, ``Movement characteristics
  of upper extremity prostheses during basic goal-directed tasks,'' {\em Clin.
  Biomech.}, vol.~25, pp.~523--529, July 2010.

\bibitem{Merad2018-ju}
M.~Merad, {\'E}.~de~Montalivet, A.~Touillet, N.~Martinet, A.~Roby-Brami, and
  N.~Jarrass{\'e}, ``Can we achieve intuitive prosthetic elbow control based on
  healthy upper limb motor strategies?,'' {\em Front. Neurorobot.}, vol.~12,
  p.~1, Feb. 2018.

\bibitem{Zhang2018-kl}
M.~Zhang, S.~Vikram, L.~Smith, P.~Abbeel, M.~J. Johnson, and S.~Levine,
  ``{SOLAR}: Deep structured representations for {Model-Based} reinforcement
  learning,'' Aug. 2018, 1808.09105.

\bibitem{Gu2016-ge}
S.~Gu, E.~Holly, T.~Lillicrap, and S.~Levine, ``Deep reinforcement learning for
  robotic manipulation with asynchronous {Off-Policy} updates,'' {\em Robotics
  and Automation}, Oct. 2016, 1610.00633.

\bibitem{Pilarski2011-nh}
P.~M. Pilarski, M.~R. Dawson, T.~Degris, F.~Fahimi, J.~P. Carey, and R.~S.
  Sutton, ``Online human training of a myoelectric prosthesis controller via
  actor-critic reinforcement learning,'' {\em IEEE Int. Conf. Rehabil. Robot.},
  vol.~2011, p.~5975338, 2011.

\bibitem{Ross2010-br}
S.~Ross, G.~J. Gordon, and J.~Andrew~Bagnell, ``A reduction of imitation
  learning and structured prediction to {No-Regret} online learning,'' Nov.
  2010, 1011.0686.

\bibitem{Gao2018-ql}
Y.~Gao, {Huazhe}, {Xu}, J.~Lin, F.~Yu, S.~Levine, and T.~Darrell,
  ``Reinforcement learning from imperfect demonstrations,'' Feb. 2018,
  1802.05313.

\bibitem{Vasan2017-og}
G.~Vasan and P.~M. Pilarski, ``Learning from demonstration: Teaching a
  myoelectric prosthesis with an intact limb via reinforcement learning,'' {\em
  IEEE Int. Conf. Rehabil. Robot.}, vol.~2017, pp.~1457--1464, July 2017.

\bibitem{Ficuciello2019-wo}
F.~Ficuciello, A.~Migliozzi, G.~Laudante, P.~Falco, and B.~Siciliano,
  ``Vision-based grasp learning of an anthropomorphic hand-arm system in a
  synergy-based control framework,'' {\em Science Robotics}, vol.~4,
  p.~eaao4900, Jan. 2019.

\bibitem{Sharif2020-ll}
M.~Sharif, D.~Erdogmus, C.~Amato, and T.~Padir, ``Towards {End-to-End} control
  of a robot prosthetic hand via reinforcement learning,'' in {\em 2020 8th
  {IEEE} {RAS/EMBS} International Conference for Biomedical Robotics and
  Biomechatronics ({BioRob})}, pp.~641--647, Nov. 2020.

\bibitem{Grosz1996-hh}
B.~J. Grosz and S.~Kraus, ``Collaborative plans for complex group action,''
  {\em Artif. Intell.}, vol.~86, pp.~269--357, Oct. 1996.

\bibitem{Nikolaidis2017-ce}
S.~Nikolaidis, Y.~X. Zhu, D.~Hsu, and S.~Srinivasa, ``{Human-Robot} mutual
  adaptation in shared autonomy,'' in {\em Proceedings of the 2017 {ACM/IEEE}
  International Conference on {Human-Robot} Interaction - {HRI} '17}, (New
  York, New York, USA), pp.~294--302, ACM Press, Jan. 2017.

\bibitem{Barredo_Arrieta2020-bw}
A.~Barredo~Arrieta, N.~D{\'\i}az-Rodr{\'\i}guez, J.~Del~Ser, A.~Bennetot,
  S.~Tabik, A.~Barbado, S.~Garcia, S.~Gil-Lopez, D.~Molina, R.~Benjamins,
  R.~Chatila, and F.~Herrera, ``Explainable artificial intelligence ({XAI)}:
  Concepts, taxonomies, opportunities and challenges toward responsible {AI},''
  {\em Inf. Fusion}, vol.~58, pp.~82--115, June 2020.

\bibitem{Amirshirzad2019-sa}
N.~Amirshirzad, A.~Kumru, and E.~Oztop, ``Human adaptation to {Human--Robot}
  shared control,'' {\em IEEE Transactions on Human-Machine Systems}, vol.~49,
  pp.~126--136, Apr. 2019.

\bibitem{Lewis2018-dq}
M.~Lewis, K.~Sycara, and P.~Walker, ``The role of trust in {Human-Robot}
  interaction,'' in {\em Foundations of Trusted Autonomy} (H.~A. Abbass,
  J.~Scholz, and D.~J. Reid, eds.), pp.~135--159, Cham: Springer International
  Publishing, 2018.

\bibitem{Levine2016-cq}
S.~Levine, P.~Pastor, A.~Krizhevsky, J.~Ibarz, and D.~Quillen, ``Learning
  {Hand-Eye} coordination for robotic grasping with deep learning and
  {Large-Scale} data collection,'' {\em The Int'l Journal of Robotics
  Research}, p.~027836491771031, Mar. 2016, 1603.02199.

\bibitem{Tassa2018-qp}
Y.~Tassa, Y.~Doron, A.~Muldal, T.~Erez, Y.~Li, D.~de~Las~Casas, D.~Budden,
  A.~Abdolmaleki, J.~Merel, A.~Lefrancq, T.~Lillicrap, and M.~Riedmiller,
  ``{DeepMind} control suite,'' Jan. 2018, 1801.00690.

\bibitem{Dam1998-zj}
E.~B. Dam, M.~Koch, and M.~Lillholm, {\em Quaternions, interpolation and
  animation}, vol.~2.
\newblock Citeseer, 1998.

\bibitem{Hoff1993-xu}
B.~Hoff and M.~A. Arbib, ``Models of trajectory formation and temporal
  interaction of reach and grasp,'' {\em J. Mot. Behav.}, vol.~25,
  pp.~175--192, Sept. 1993.

\bibitem{Haarnoja2018-ts}
T.~Haarnoja, A.~Zhou, P.~Abbeel, and S.~Levine, ``Soft {Actor-Critic}:
  {Off-Policy} maximum entropy deep reinforcement learning with a stochastic
  actor,'' Jan. 2018, 1801.01290.

\bibitem{Kingma2014-fb}
D.~P. Kingma and J.~Ba, ``Adam: A method for stochastic optimization,'' Dec.
  2014, 1412.6980.

\bibitem{Borges2018-aa}
M.~Borges, A.~Symington, B.~Coltin, T.~Smith, and R.~Ventura, ``{HTC} vive:
  Analysis and accuracy improvement,'' in {\em 2018 {IEEE/RSJ} International
  Conference on Intelligent Robots and Systems ({IROS})}, pp.~2610--2615, Oct.
  2018.

\bibitem{Olson2011-zk}
E.~Olson, ``{{AprilTag}}: A robust and flexible visual fiducial system,'' in
  {\em 2011 {IEEE} International Conference on Robotics and Automation},
  pp.~3400--3407, May 2011.

\bibitem{Sharif2019-ee}
M.~Sharif, ``Digideep: A {DeepRL} pipeline for developers,'' 2019.

\bibitem{Paszke2019-ty}
A.~Paszke, S.~Gross, F.~Massa, A.~Lerer, J.~Bradbury, G.~Chanan, T.~Killeen,
  Z.~Lin, N.~Gimelshein, L.~Antiga, A.~Desmaison, A.~Kopf, E.~Yang, Z.~DeVito,
  M.~Raison, A.~Tejani, S.~Chilamkurthy, B.~Steiner, L.~Fang, J.~Bai, and
  S.~Chintala, ``{PyTorch}: An imperative style, {High-Performance} deep
  learning library,'' in {\em Advances in Neural Information Processing Systems
  32} (H.~Wallach, H.~Larochelle, A.~Beygelzimer, F.~d'Alch{\'e} Buc, E.~Fox,
  and R.~Garnett, eds.), pp.~8024--8035, Curran Associates, Inc., 2019.

\bibitem{Todorov2012-ty}
E.~Todorov, T.~Erez, and Y.~Tassa, ``{MuJoCo}: A physics engine for model-based
  control,'' in {\em 2012 {IEEE/RSJ} International Conference on Intelligent
  Robots and Systems}, pp.~5026--5033, Oct. 2012.

\bibitem{Kumar2015-cm}
V.~Kumar and E.~Todorov, ``{MuJoCo} {HAPTIX}: A virtual reality system for hand
  manipulation,'' in {\em 2015 {IEEE-RAS} 15th International Conference on
  Humanoid Robots (Humanoids)}, pp.~657--663, Nov. 2015.

\bibitem{Medynski2011-qg}
C.~Medynski and B.~Rattray, ``Bebionic prosthetic design,''
  dukespace.lib.duke.edu, 2011.

\bibitem{Sadeghi2016-xt}
F.~Sadeghi and S.~Levine, ``{CAD2RL}: Real {Single-Image} flight without a
  single real image,'' Nov. 2016, 1611.04201.

\bibitem{Tobin2017-ew}
J.~Tobin, R.~Fong, A.~Ray, J.~Schneider, W.~Zaremba, and P.~Abbeel, ``Domain
  randomization for transferring deep neural networks from simulation to the
  real world,'' Mar. 2017, 1703.06907.

\end{thebibliography}

\end{document}